\documentclass[letterpaper, 10 pt, journal, twoside]{IEEEtran}

\usepackage[utf8]{inputenc}
\usepackage{hyperref}
\usepackage{amsmath}
\usepackage{amssymb}
\usepackage{graphicx}
\usepackage{dsfont}
\usepackage{todonotes}
\usepackage[T1]{fontenc}
\usepackage{textcomp}
\usepackage{subfig}

\newcommand{\etal}{\textit{et al}. }

\begin{document}

\title{Mine Tunnel Exploration using Multiple Quadrupedal Robots}

\author{Ian D. Miller$^{1}$, Fernando Cladera$^{1}$, Anthony Cowley$^{1}$, Shreyas S. Shivakumar$^{1}$, Elijah S. Lee$^{1}$, \\
Laura Jarin-Lipschitz$^{1}$, Akhilesh Bhat$^{1}$, Neil Rodrigues$^{1}$, Alex Zhou$^{1}$, Avraham Cohen$^{1}$, \\
Adarsh Kulkarni$^{1}$, James Laney$^{2}$,
Camillo Jose Taylor$^{1}$, and Vijay Kumar$^{1}$ 
\thanks{Manuscript received: September 10, 2019; Revised December 24, 2019; Accepted January 3, 2020.}
\thanks{This paper was recommended for publication by Editor Jonathan Roberts upon evaluation of the Associate Editor and Reviewers' comments.
This work was supported by the MAST Collaborative Technology Alliance -
Contract No. W911NF-08-2-0004, ARL grant W911NF-08-2-0004, ONR
grants N00014-07-1-0829, N00014-14-1-0510, ARO grant W911NF-13-1-
0350, NSF grants IIS-1426840, IIS-1138847, DARPA grants HR001151626,
HR0011516850, and in part by the Semiconductor Research Corporation (SRC),
DARPA, and a NASA Space Technology Research Fellowship.}
\thanks{$^{1}$The authors are with the GRASP Laboratory,
        University of Pennsylvania, Philadelphia PA 19104. 
        {\tt\footnotesize  iandm@seas.upenn.edu}} 
\thanks{$^{2}$James Laney is with Ghost Robotics,
        Philadelphia PA 19146.} 
\thanks{Digital Object Identifier (DOI): see top of this page.}
}


\maketitle

\begin{abstract}
Robotic exploration of underground environments is a particularly challenging problem due to communication, endurance, and traversability constraints which necessitate high degrees of autonomy and agility.  These challenges are further exacerbated by the need to minimize human intervention for practical applications.  While legged robots have the ability to traverse extremely challenging terrain, they also engender new challenges for planning, estimation, and control.

In this work, we describe a fully autonomous system for multi-robot mine exploration and mapping using legged quadrupeds, as well as a distributed database mesh networking system for reporting data.  In addition, we show results from the DARPA Subterranean Challenge (SubT) Tunnel Circuit demonstrating localization of artifacts after traversals of hundreds of meters.  These experiments describe fully autonomous exploration of an unknown Global Navigation Satellite System (GNSS)-denied environment undertaken by legged robots.
\end{abstract}


\section{Introduction}

\IEEEPARstart{A}{} long-awaited promise of mobile robotics, almost from its inception, has been the autonomous exploration of environments inhospitable to humans.  Since at least the early 2000s, work has been ongoing to autonomously explore the particularly dangerous and complex environment of subterranean mines \cite{Thrun2004} \cite{Morris2005}.  In this work, we consider an ``autonomous mission'' to be a mission where there is no human interaction after mission start.  
Mines present a wide variety of challenges to virtually all aspects of robotics, including challenging and complex terrain, inherent GNSS and communication denial, and sensory degradation.  In addition, if robots are to be used practically in the field, procedures for robot deployment and operator control must be highly automated, robust, and reliable.

In an effort to accelerate development of technologies to addresss these challenges, DARPA are running the Subterranean Challenge (SubT).  The goal of SubT is to detect, localize, and report as many artifacts as possible in an hour over a course of multiple kilometers of tunnel.  Artifact categories are known \textit{a priori} and include objects such as backpacks, manikins, and cellphones.  Systems are not required to be autonomous, but all direct robot communication must be handled by one person, and the environment naturally imposes severe communication constraints \cite{Yarkan2009}.

\begin{figure}
    \centering
    \includegraphics[width=0.9\linewidth]{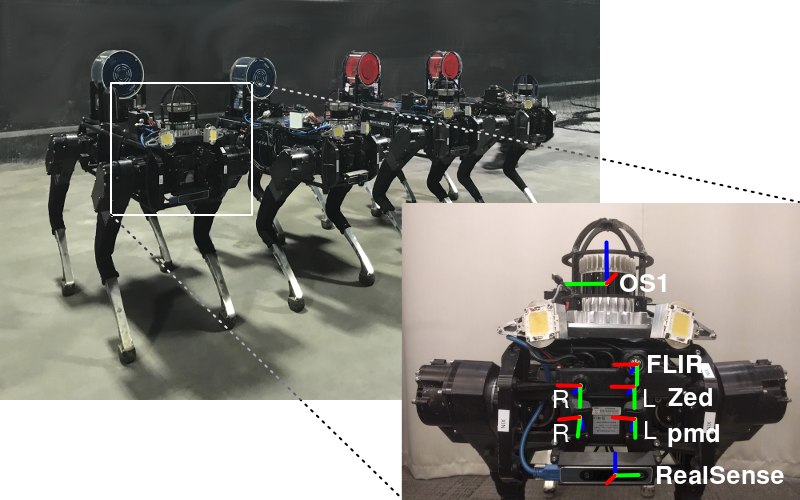}
    \caption{Quadrupedal platforms used at Tunnel Circuit with expanded view of sensor suite.}
    \vspace{-2em}
    \label{fig:robots}
\end{figure}

Limited bandwidth and connectivity require robotic systems to operate successfully without any real-time feedback from the human operator.  All subsystems must run entirely on-board on relatively compute-limited platforms.  Furthermore, significant attention must be given to data compression, transmission, and storage in order to provide the human operator with the most useful information over a limited-bandwidth link.

In addition, underground environments contain a wide variety of terrains (e.g.~concrete, deep mud, pitted railroad ties, and vertical shafts) over kilometers of tunnel.  Since no single platform, or type of platform, is perfectly suited to all of these environments, we propose a loosely-coupled heterogeneous multi-robot system consisting of Micro-Aerial Vehicles (MAVs) and legged quadrupeds.  Recent work has noted the capabilities for MAVs in underground environments to traverse areas rapidly that ground-based systems cannot reach \cite{Papachristos2019} \cite{Li2018} \cite{Ozaslan2018}. However, there are limits on the flight time, mission length, and payload that such platforms can support.  We therefore additionally employ Ghost Robotics (GR) Vision 60 quadrupeds (Fig.~\ref{fig:robots}) to enable long duration missions with a larger sensor payload.  Quadrupeds have the potential capability to traverse highly complex and unstructured terrain such as steps more easily than treaded or wheeled robots, making them suited for underground environments \cite{Zimroz2019}, but they have been minimally utilized in real-world autonomous applications.  In a work similar to ours, Bellicoso \etal \cite{Bellicoso2018} navigate outdoor environments with a quadruped, but they (a) assume GNSS availability and (b) assume a pre-existing global map or manual path definition.

In this work, we discuss the system design for the legged robots, but note that the detection and communication portions of our approach are platform-agnostic and are utilized across a heterogeneous suite of robots including quadrotors.  

The primary contributions of our work are as follows:
\begin{itemize}
    \item We present the architecture and detail the components of our exploration and planning algorithms for autonomous exploration of underground tunnel environments.
    \item We discuss communication, detection, and mapping systems that enable rapid situational awareness for a single operator in  communication-challenged environments.
    \item We show results from testing in a motion capture space in the lab, as well as the National Institute for Occupational Safety and Health (NIOSH) mine at the SubT Tunnel Circuit, including multiple autonomous traversals of hundreds of meters in different mines.  These experiments describe long-duration entirely autonomous exploration undertaken by legged robots.
\end{itemize}
\section{System Overview}

\subsection{Hardware Architecture}

 We chose the Nvidia Jetson AGX Xavier as the primary computer for the legged platform.  The Xavier contains 8 ARM CPU cores as well as 512 GPU CUDA cores with 16GB 
 of shared RAM.  We can therefore distribute the software load between the CPU and GPU, which is particularly valuable for tasks such as mapping and image processing.

\begin{figure}
    \centering
    \vspace{2mm}
    \includegraphics[width=0.9\linewidth]{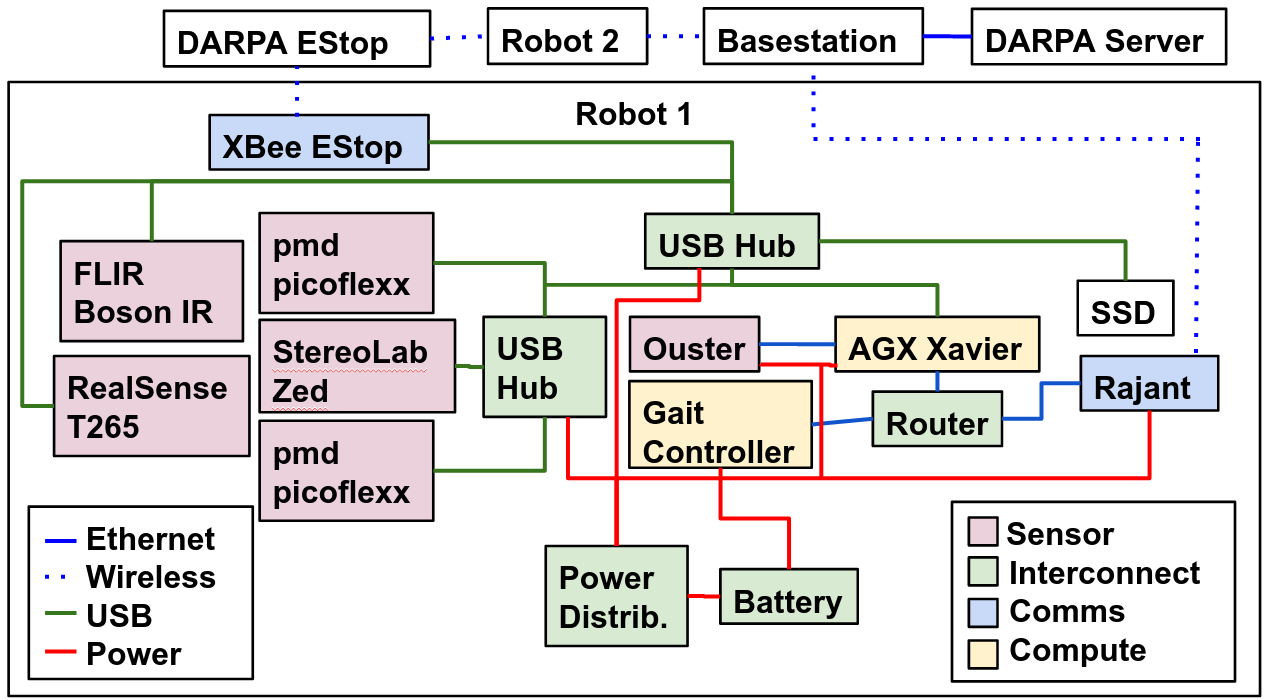}
    \caption{Hardware architecture of individual robot showing sensors and compute as well as interconnect to other robots, the basestation, and the DARPA infrastructure.}
    \vspace{-2em}
    \label{fig:hw_block}
\end{figure}

The hardware architecture is shown in Fig.~\ref{fig:hw_block}.  Our primary exteroceptive sensor is an Ouster OS-1 64-beam LiDAR, which is used for navigation, planning, and mapping.  Two pmd picoflexx time-of-flight sensors are used for short-range depth sensing in front of the robot to aid in local planning.  We employ an Intel RealSense T265 stereo tracking camera for local high-speed visual-inertial pose estimation. A StereoLabs Zed Mini stereo camera, FLIR Boson IR camera, and Bluetooth radio are the primary sensors used for object detection, with the thermal camera primarily useful for detecting survivors (humans) due to body heat and the bluetooth detector for detecting cell phones.  An XBee is required for emergency stop override capability by DARPA.

For communication, we use a Rajant dual-band mesh networking system.  The Xavier additionally communicates with the Vision 60 gait controller via UDP over Ethernet to send high-level navigation commands in the form of body-frame twists.

\subsection{Software Architecture}

Our full software stack is built on the Robot Operating System (ROS) middleware.  Because communication is intermittent and unreliable in subterranean environments, all robots and external computers run separate ROS masters.  Communication between robots occurs exclusively via the distributed database system (Sec.~\ref{sec:dist_db}), and communication with the basestation via a combination of the distributed database and a separate mechanism for sending commands and obtaining telemetry.  Subsystems, packaged as ROS nodes, are launched, stopped, and monitored by a launch manager daemon.  In this fashion, the operator can individually stop and start the robot's subsystems, as well as verify that each is operating as expected without needing an SSH or direct ROS connection to the robot.  This interface is pictured in Fig.~\ref{fig:top_down}.  This approach proved to be extremely useful for rapid startup, shutdown, and debug of systems in the field.

\begin{figure}
    \centering
    \vspace{2mm}
    \includegraphics[width=0.87\linewidth]{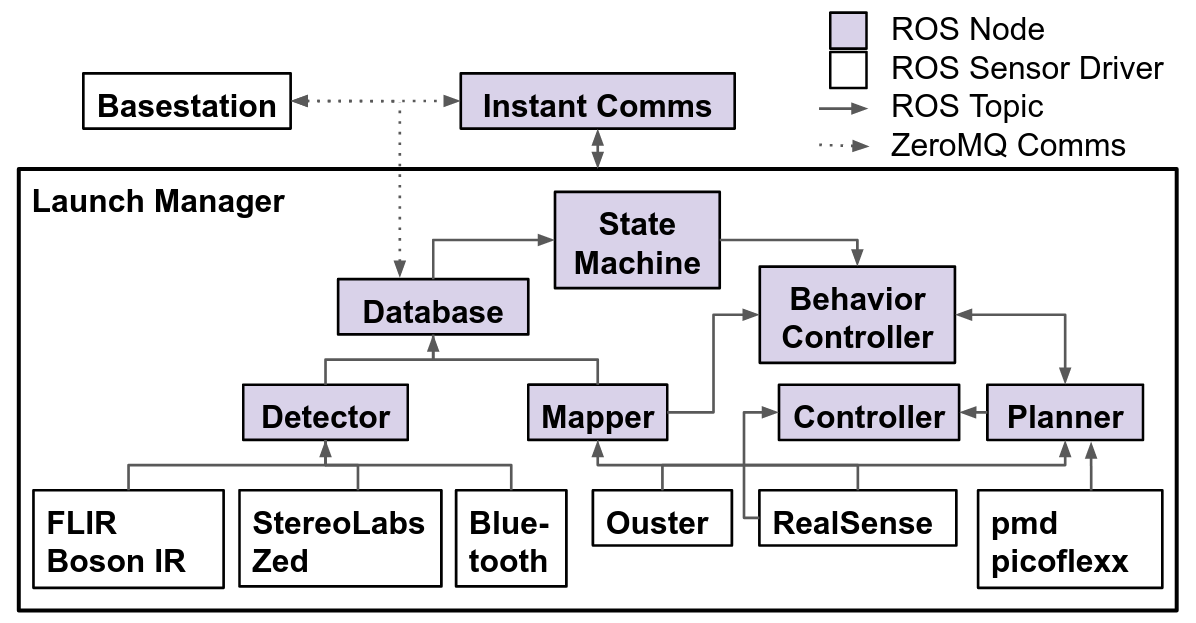}
    \caption{High level software architecture.  Aside from the basestation, all ROS nodes run locally on the robot.  The launch manager handles starting and stopping its' member nodes based on commands sent through the basestation.}
    \vspace{-1.5em}
    \label{fig:soft_block}
\end{figure}

A block diagram of the software architecture is shown in Fig.~\ref{fig:soft_block}.  Each subsystem is largely independent, enabling a high degree of robustness.  We made an early decision to separate the lower level autonomy systems, such as the local planner and controller, from higher level systems such as the mapper and state machine.  Therefore, even if the global mapper fails or diverges, the robot remains capable of locally determining feasible directions to travel. This robustness was found to be very useful in experiments.

\section{Mapping} \label{mapping}

\begin{figure}
    \centering
    \includegraphics[width=\linewidth]{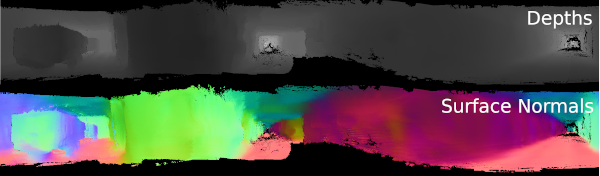}
    \caption{Local map keyframes include depth and surface normal information}
    \label{fig:mappper_keyframe}
    \vspace{-2em}
\end{figure}

The mapper is based on an atlas framework \cite{Bosse03anatlas} with a pose graph \cite{Dellaert2012FactorGA} skeleton linking together local maps, or keyframes. The local maps support small-scale loop closures by virtue of powering local frame-to-model tracking in the spirit of KinectFusion \cite{Newcombe:KinectFusion:2011}, while the pose graph enables large-scale loop closures. The local map representation is a set of panoramas \cite{Meilland13b,Taylor:DepthPanoramas:2015} recording depth, surface normal, and confidence for each pixel (Fig.~\ref{fig:mappper_keyframe}). Panoramic images efficiently represent free space, as it is implicitly encoded along the ray to each recorded range measurement, without sacrificing much representational power when used for maps gathered over limited translational excursions.

We set the LiDAR scan frequency at 10Hz. Robot motion over these 100ms intervals can be significant, so an initial rotation correction based on a time-synchronized IMU is used to ``de-rotate'' measurements to the time when the sweep began. The rigid motion of each sweep is initialized by the pose change reported by an Intel RealSense stereo camera maintaining its own independent pose estimate, then optimized to register the sweep to the active keyframe with projective iterative closest point (ICP) \cite{Blais:ProjectiveICP:1995} with a point-to-plane metric \cite{Chen:ICP:1992}. The registered sweep is used to update the active keyframe by averaging range estimates that are within a small threshold, or decreasing the confidence of an estimate with which the new observation disagrees. A new keyframe is initialized from the previous keyframe whenever the current map can not represent enough of a new sweep due to occlusions or changes in the environment, or if the current pose is above a threshold distance from the keyframe origin.

The mapping pipeline typically runs at 50Hz on the Nvidia Xavier, with most of the computation handled by the GPU. When a new keyframe is created, a pose graph including the just-finished active map is optimized using GTSAM. The depth image for the newly finished keyframe is then downsampled and losslessly compressed as a 16-bit PNG image for transmission back to the base station for use in generating a traditional point cloud visualization. The base station's global point cloud is assembled from individual robot keyframes into a global pointcloud.
\section {High Level Navigation} 

Due to our decision to decouple exploration from mapping, our exploration algorithm was developed to operate entirely locally.  A tunnel exploration algorithm must be able to both determine potential tunnels for exploration, as well as decide which tunnel at any given time to explore.

\subsection {Tunnel Detection} \label{sec:tunnel_detection}

\begin{figure}
    \centering
    \vspace{2mm}
    \includegraphics[width=\linewidth]{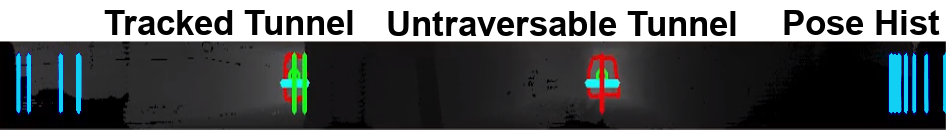}
    \caption{Depth Panorama with currently tracked tunnels and projected pose history.}
    \vspace{-1em}
    \label{fig:depth_pano}
\end{figure}

It is natural to represent tunnels at a high level by a graph as in \cite{Morris2005}, as the confined nature of the environment results in high degrees of sparsity.  For exploration, detection of crossroads is of particular significance \cite{Silver2006} as they represent decision points.  We use instantaneous depth panoramas generated by a single 3D LiDAR scan as the primary input to our exploration algorithm due to several useful properties.  A large section of distant points, corresponding to a connected component in the depth panorama, represents a tunnel.  Therefore, we threshold the depth panorama at a target distance, detect connected components of sufficient area, and find the corresponding centroids.  These centroids represent the azimuths of potential tunnel axes emanating from the robot's current position.  We then track these centroids over time in the LiDAR frame using an EKF, associating centroid measurements to tracked tunnels by checking whether the detected centroid is within the tunnel bearing uncertainty maintained by the EKF.  If this uncertainty drops below a threshold, the robot is considered to have detected a new tunnel.  In order to avoid inadvertent backtracking, we track the recent pose history and project these bearings onto the depth panorama, ignoring centroids near them.  This can be seen in Fig.~\ref{fig:depth_pano}.

\subsection {Exploration Behavior} \label{exp_beh}

We continuously track the tunnel closest to the robot's current heading.  If the local terrain planner informs the exploration controller that the cost of traversal is too high, as described in Sec.~\ref{terrain_mapping}, then the tunnel is marked as untraversable and the next closest tunnel selected.  If there are no visible tunnels, a return to base is triggered.

In order to ensure multiple robots take different trajectories, we expose simple mission specifications to the operator in the form of turn sequences such as ``Left $\rightarrow$ Right $\rightarrow$ Right''.  The operator inputs the desired specification manually at the beginning of the mission, generally seeking to give different robots different specifications in order to avoid redundant exploration.
The robot queues up these turns, and if a new tunnel is detected (see Sec.~\ref{sec:tunnel_detection}) on the relevant side of the robot, that one is automatically selected as the new current tunnel and the turn command is popped off the top of the queue.  Once the queue is emptied, the robot simply continues along the current tunnel until it becomes incapable of continuing or exploration times out.

\subsection {Return to Base Behavior}
In order to return to the entrance to the tunnel we leverage the mapper's topological graph as described in Sec.~\ref{mapping}. By performing a graph search, we can find a path from the current keyframe to the start keyframe. Since keyframes are typically evenly spaced, we traverse the graph using breadth-first search. Once a path is found, we compute a vector from our current position estimate to the next vertex in the path (which we subsequently update as we get nearby). This vector is fed to our existing planning framework of tunnel following. When the direction is similar to the direction of a detected tunnel, we follow the tunnel as in section \ref{exp_beh}. When it is above the angular difference threshold, we ignore detected tunnels and plan paths in that direction (as in the case of turning around at the beginning of the path). This `go to keyframe' behavior could be used without modification to navigate to other keyframes in the pose graph, such as keyframes adjacent to unexplored tunnels, but in practice was used only to return to the start keyframe.
\section {Local Planning and Control}

\begin{figure}
    \centering
    \vspace{2mm}
    \includegraphics[width=0.38\textwidth]{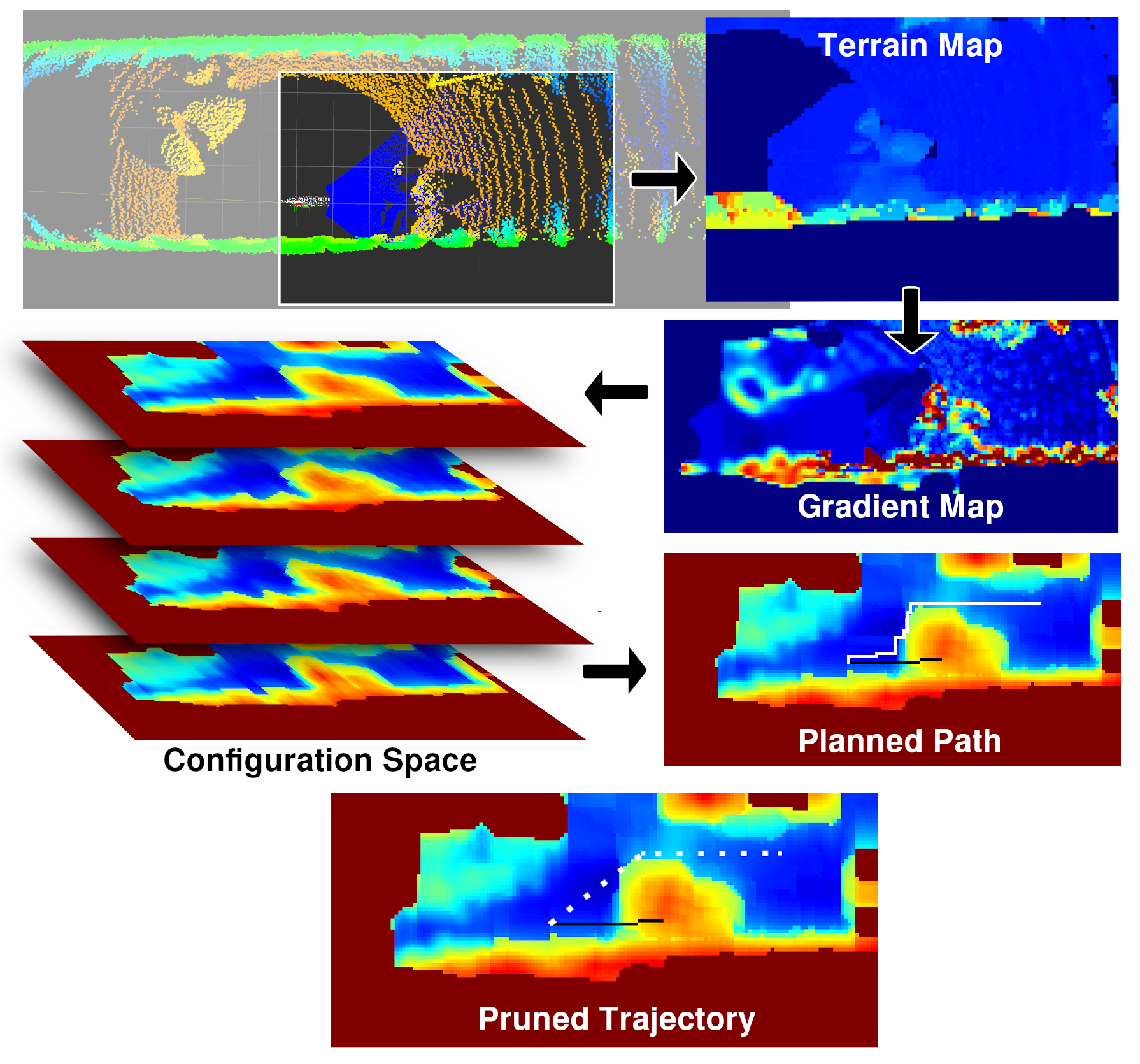}
    \caption{Planning pipeline from pointclouds to final pruned trajectory}
    \vspace{-1em}
    \label{fig:planner_pipeline}
\end{figure}

It is common, when considering the motion planning problem for ground robots, to encode the local environment as a 2-dimensional costmap as in \cite{Gerkey2008}, where the cost is effectively the inverse of some measure of traversability.  The problem of local planning is then broken into two primary pieces: quantifying terrain traversability and planning feasible trajectories on the generated cost map.  Due to imperfect odometry and our decision to minimally couple planning and mapping, we adopt the approach of \cite{Fankhauser2018} and continually center the costmap on the robot's current position.  We additionally model traversability as a function of the magnitude of the terrain gradient, which makes costmap integration extremely simple.  The full planning pipeline is shown in Fig.~\ref{fig:planner_pipeline}.

\subsection {Terrain Characterization} \label{terrain_mapping}

We begin by converting each LiDAR scan to a discretized heightmap.  Ceilings are filtered out, and in each grid cell on the ground (X-Y) plane, we compute the average height.  Once we have a local heightmap, it is trivial to compute its gradient.  We motivate this operation by noting that for ground robots, the traversability of the terrain is largely determined by steepness and roughness, both of which are captured in the gradient.  We consider only the magnitude of the gradient at each point.  Additionally, we make the conservative assumption that regions with unknown gradient have infinite cost.

Finally, by using the gradient the absolute elevation is effectively filtered out.  We can therefore make the assumption that all robot motion is in the 2D X-Y ground plane when integrating without significant error.  Let $T_2(t)$ be the robot pose projected into SE(2) at time $t$.  Let the maintained local gradient map be $G(t)$, and the computed map from a new LiDAR scan be $G_m(t)$.  We index the cells on the X-Y plane by $(x,y)$.  We then have
\begin{equation}
    G(t+1) = T_2(t+1) T_2^{-1}(t) G(t)
\end{equation}
We then fuse the propagated old map and the new measured map $G_m(t+1)$ with
\begin{equation}
\begin{split}
    &G(t+1)_{x,y} = \\
    &\begin{cases}
        G_m(t+1)_{x,y} & G(t+1)_{x,y} = \text{NaN} \\
        (G_m(t+1)_{x,y} + G(t+1)_{x,y})/2 & G(t+1)_{x,y} \neq \text{NaN} \\
        & G_m(t+1)_{x,y} \neq \text{NaN} \\
        \text{NaN} & \text{else}
    \end{cases}
\end{split}
\end{equation}

We now convert the gradient map in Cartesian space into configuration space $C$ by modelling the robot as a 2D rectangular region $R(x,y,\theta)$ centered on $(x,y)$ with angle $\theta$.  We then define
\begin{equation}
    C_{x,y,\theta} = \text{max}\{g_{x,y} \in G \; | \; (x,y) \in R(x,y,\theta) \}
\end{equation}
which is the most conservative possible traversability assumption.  If any $g_{x,y} \in G$ are unknown (NaN), we set $C_{x,y,\theta} = \infty$, again making the conservative assumption that all unknown space is untraversable.  The configuration space conversion lends itself to parallelization, so we perform this step on the GPU.  We additionally precompute the bounds of $R(x,y,\theta)$ in a lookup table, so the entire configuration space generation takes \textasciitilde 10 ms on the Xavier.  As a final step, we blur the configuration space using a 1D Gaussian kernel along the y axis.  This increases the cost of cells near the tunnel walls on either side of the robot, pushing the robot towards the center of the tunnel.

\subsection {Terrain Planning}


A common planning approach for legged robots is to plan a sequence of footstep placements on the traversability map such as in \cite{Kalakrishnan2010} such that the placements are feasible and stable.  Such approaches, while capable of navigating extremely complex terrain, are also computationally intensive as trajectory optimization must be performed for every step.  We adopt an approach similar to \cite{Wermelinger2016}, where we plan based on the traversability of a \textit{robot} "footprint" in configuration space and do not plan individual steps.  The cost of a trajectory is then the sum of a number of penalties for traversability and trajectory length.  Our method differs from theirs in that we do not have an explicit goal state, but rather a general goal direction.  This method is particularly apropos for legged robots which have less of a binary understanding of traversable/not traversable and more of a continuous spectrum of probability of traversability.

The planner can be defined to be a function $P(C) : \mathbf{d} \rightarrow \tau'$ where $\mathbf{d} \in \mathbb{R}^2$ is a 2D direction and $\tau' = [\tau'(t_1), \tau'(t_2), ..., \tau'(t_n)]$, $\tau'(t_i) \in SE(2)$ is the optimal time parametrized trajectory.  We additionally define
\begin{equation}
    \tau' = \text{argmin}_\tau\{L(\tau) + D(\tau) \; | \; \tau \in \mathbb{T} \}
\end{equation}
where $L(\tau)$ is the cost function, $D(\tau)$ is the cost associated with commanded direction, and $\mathbb{T}$ is the set of admissible trajectories.  This hybrid planner allows us to impose both "hard" and "soft" constraints on $\tau$.  We define $\mathbb{T}$ to be the set of trajectories which never go through a cell in configuration space of cost greater than a threshold.

We define
\begin{equation}
\begin{split}
    L(\tau) =
    \sum_{t=0}^T [&L_\theta (\tau_\theta(t+\delta t) - \tau_\theta(t)) + \\
                  &L_{x,y} ||\delta \tau_{x,y}(t)||_2 + \\
                  &L_{G} \log(C(\tau(t))||\delta \tau_{x,y}(t)||_2) + \\
                  &L_{s} (\arctan(\frac{\tau_y(t+\delta t) - \tau_y(t)}{\tau_x(t+\delta t) - \tau_x(t)}) - \tau_\theta(t))^2 + \\
                  &L_{b} \max(-\delta \tau_{x,y}(t) \cdot [\cos{\tau_\theta(t)}, \sin{\tau_\theta(t)}]), 0)]
\end{split}
\end{equation}
where $\delta \tau_{x,y}(t) = \tau_{x,y}(t+\delta t) - \tau_{x,y}(t)$.  These terms correspond to a rotation cost, distance cost, traversability cost, sidestep cost, and reversal cost.  Notably, we additionally penalize sidestepping on the robot as it is significantly less stable while sidestepping when compared to moving directly forward.  We strongly penalize backwards motion since no sensors on the robot can instantaneously see backwards.  In order to direct the path in the desired direction, we add the direction cost $D(\tau) = L_D (||\tau_{x,y}(T) - \tau_{x,y}(0)||_2 - (\tau_{x,y}(T) - \tau_{x,y}(0)) \cdot d_{\text{des}})$ where $d_{\text{des}}$ is a unit vector in the desired direction.  Notably, this approach assumes that the high level exploration system is generating local directions which are generally open, an assumption we found to be typically true, though the high level tunnel tracker will pick a new direction if the path cost is too high.

We run Dijkstra's algorithm over configuration space to plan the initial path, terminating the search once we have reached any point sufficiently far along $d_\text{des}$.  Once the path is found, we recursively prune the planned path to remove all waypoints that do not increase the overall cost and are in $\mathbb{T}$, linearly interpolating between waypoints.  Pruning smooths the final trajectory, removing the jagged turns that are an artifact of planning in discretized Cartesian space.

\subsection {Control}

Once the planned path has been generated, we time parametrize it assuming fixed linear and angular velocities.  At every time $t$ we then compute $\tau(t)$ as well as $x(t)$, the current actual state of the planner.  We finally simple use a simple proportional controller to command a twist proportional to $\tau(t) - x(t)$ in order to track the trajectory.

\section{Object Detection}

\subsection {Vision Based Detection}

\begin{figure}
    \centering
    \vspace{3mm}
    \includegraphics[width=\linewidth]{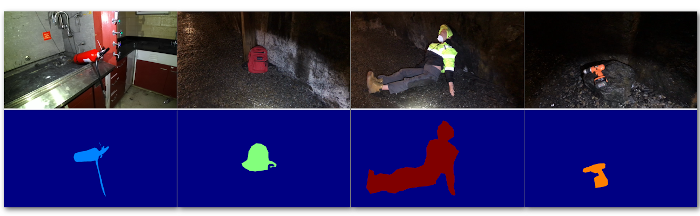}
    \caption{Examples of artifacts with per-pixel annotations for training data}
    \vspace{-1.5em}
    \label{fig:detection_data}
\end{figure}

The artifact types for the SubT Tunnel Circuit are fire-extinguisher, backpack, hand-drill, cellphone and survivor; examples are shown in Fig.~\ref{fig:detection_data}. We achieve artifact recognition using the ERFNet \cite{romera2017erfnet} deep learning-based semantic segmentation network. We selected this network architecture because of its compact and efficient design, allowing fast inference on our compute hardware \cite{MAVNet}. 

We collected a large corpus of training data from various challenging environments such as mines, basements, and cluttered indoor and outdoor spaces from a variety of sensors including three different types of cellphones and Stereolabs Zed Mini cameras. We also collected human annotated per-pixel labels for each of these images. Additionally, to add more variation to our dataset, we extracted relevant image frames from over 25 YouTube videos of caves and mines. To a fraction of these images, we programmatically and artificially pasted artifacts into these scenes and automatically generated labels. This resulted in an annotated dataset of 3416 images of resolution $1280\times720$. 

For performance reasons we trained our network at half the original resolution using the dataset described above, splitting our dataset such that training and test data are from different environments. We then selected the model with the lowest test accuracy (mean intersection over union) over the relevant classes. 

The output of our semantic segmentation network, along with the depth image acquired from the Stereolabs Zed Mini passive stereo camera is used to identify the artifact's 3D position relative to the camera frame. These 3D positions are then transformed to the nearest keyframe from our mapping algorithm. 

Our segmentation network runs at \textasciitilde 5Hz and we usually receive multiple detections of the same artifact. To filter these artifact detections and reduce the number communicated back to the base-station, we cluster multiple detections based on their 3D positions relative to the last keyframe. Artifact detections of the same class that are detected within some distance threshold are collapsed to the same artifact instance and are returned to the human operator
with a high confidence score. This confidence value is calculated as the ratio of the number of detections that are in agreement with our threshold to the total number of detections within a particular time-window. However, artifacts that exist as outliers, i.e those with higher error in calculated 3D position or false positives from the neural network are also returned to the operator for final decision making.

\subsection {Cellphone Detection}

The bluetooth detector consisted of a HC-05 bluetooth module
on a FT232RL FTDI USB to TTL Serial adapter. 
The HC-05 is used in \textit{command} mode to scan for bluetooth enabled devices, as all cell phones had bluetooth on at the competition. Identified devices are then logged by their unique MAC ID and strength of bluetooth signal using the RSSI values returned during the scan.  Detections are pushed to the database each time a stronger signal corresponding to a given device was detected, tagged with the current pose of the robot.
\section{Distributed Database}%
\label{sec:dist_db}


Communication presents a key challenge in underground environments with
an intermittent connection between nodes.
Furthermore, the
information generated by one robot may be useful for other robots
deployed in the field.
Therefore, we opted to use a distributed network and data sharing
architecture, in which all
nodes share the same information
We refer
to this module as the \emph{distributed database}.
Similar
architectures have been proposed in the literature to coordinate
distributed autonomous agents
\cite{cowley2004distributed} \cite{ferrer2018blockchain}.




Each message stored in the database is identified by a unique hash,
calculated when the message is recorded.
Data synchronization between pairs of nodes is triggered automatically when the link quality between nodes is sufficiently high.  When synchronization is triggered, hashes are compared, and the messages corresponding to new or different hashes are transmitted one-by-one.  Therefore, if a transmission is interrupted, the system need only finish synchronizing the remaining messages.

While we synchronize the database between all the nodes in the network,
each node
only has read/write access to its own
data, and read-only access to the data published by other nodes. Therefore, a robot may use the information collected by
other robots to modify its
behavior without requiring complex synchronization procedures to ensure
data consistency.

\section{Strategy, Human Interface}%
\label{sec:strategy}

Given that communication is challenging, low-bandwidth, and intermittent in subterranean environments, we design our system to be totally distributed so each robot behaves exclusively based on its own observations and information. Simply put, we view communication as a luxury and focus on developing fully autonomous systems whose behaviors are triggered by individual high-level state machines. 


\subsection {State Machine}
\begin{figure}
    \centering
    \vspace{2mm}
    \includegraphics[width=0.9\linewidth]{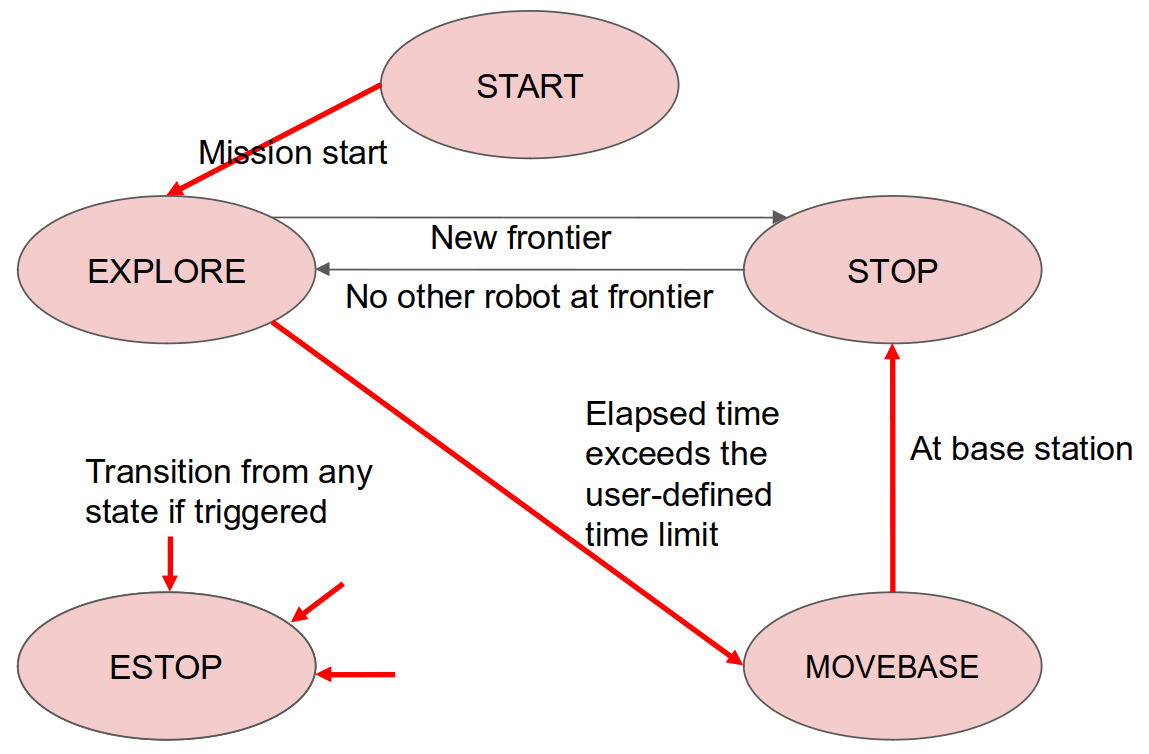}
    \caption{State machine overview.}
    \vspace{-1em}
    \label{fig:statemachine}
\end{figure}

We define multiple states for individual robots, namely \texttt{START}, \texttt{EXPLORE}, \texttt{STOP}, \texttt{MOVEBASE}, and \texttt{ESTOP}, as shown in Fig.~\ref{fig:statemachine}.  In the \texttt{START} state the operator can set the time limit for robot exploration before returning as well as the turn specification (Sec.~\ref{exp_beh}). This time limit can additionally be changed as long as the robot can communicate with the base station. When the operator commands, the robot enters \texttt{EXPLORE} and the robot actively searches for tunnels to follow, as detailed in Sec.~\ref{exp_beh}. The robot returns to the base when the elapsed time exceeds the user-defined time limit, and \texttt{MOVEBASE} is entered. \texttt{STOP} and \texttt{ESTOP} can be manually triggered to either simply pause or sit down the robot. For Tunnel Circuit, we deployed robots one-by-one with decreasing time limits in order to mitigate inter-robot interference and maintain communication through the mesh network.  Extending this system by allowing inter-robot collaboration could improve performance and remains as a future work. 

\subsection {User Interface}
\begin{figure}
    \centering
    \includegraphics[width=0.8\linewidth]{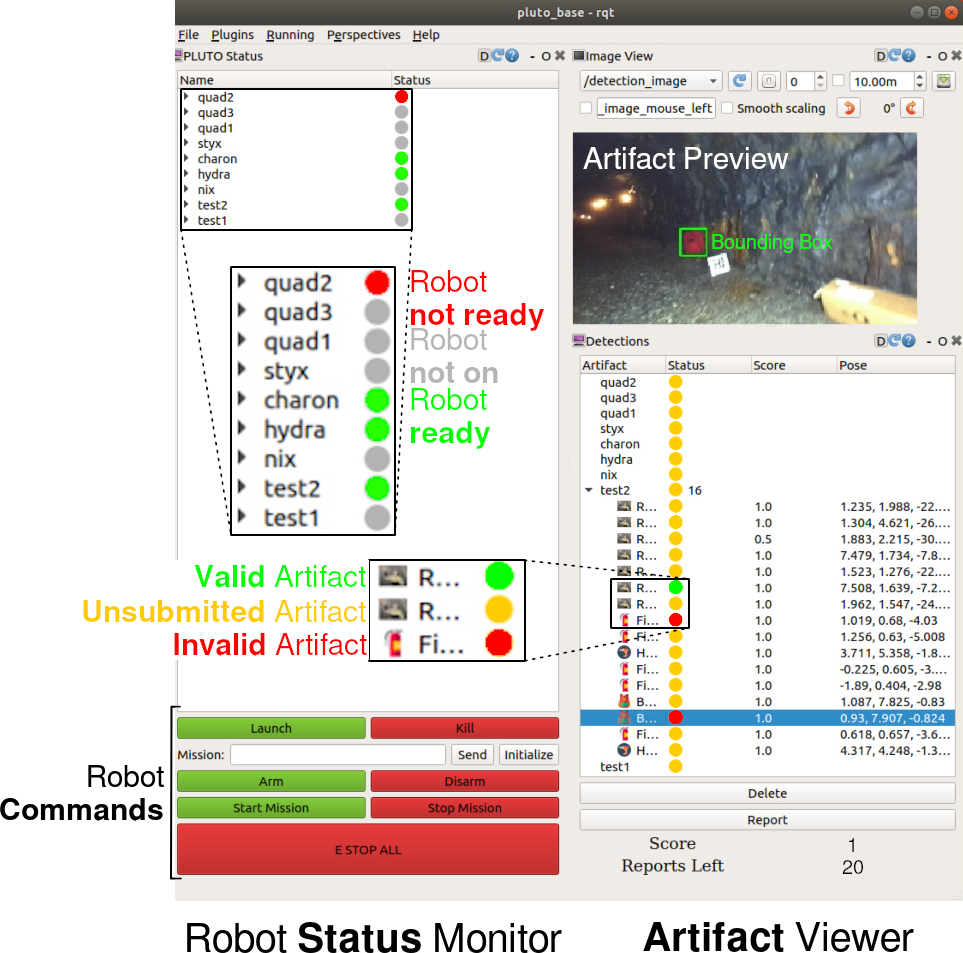}
    \caption{An instance of the user interface including the robot status monitor as well as the artifact listing from each robot and artifact viewer.}
    \vspace{-1em}
    \label{fig:top_down}
\end{figure}

To monitor the states of the robots and artifact positions, we develop a user interface as shown in Fig.~\ref{fig:top_down}.  The interface shows the individual robot status from the launch manager, the list of detected artifacts, and images with bounding boxes of detected artifacts for the operator to review before sending to DARPA.  The operator can also view a pointcloud reconstruction of the global map as well as a top-down view of the paths of each robot and the locations of detected artifacts, similar to Figs.~\ref{fig:exp_mine} and \ref{fig:sr_mine}.

\section{Experimental Results}

\subsection{Lab Experiment}

\begin{figure*}[ht]
    \centering
    \vspace{2mm}
    \includegraphics[width=0.75\textwidth]{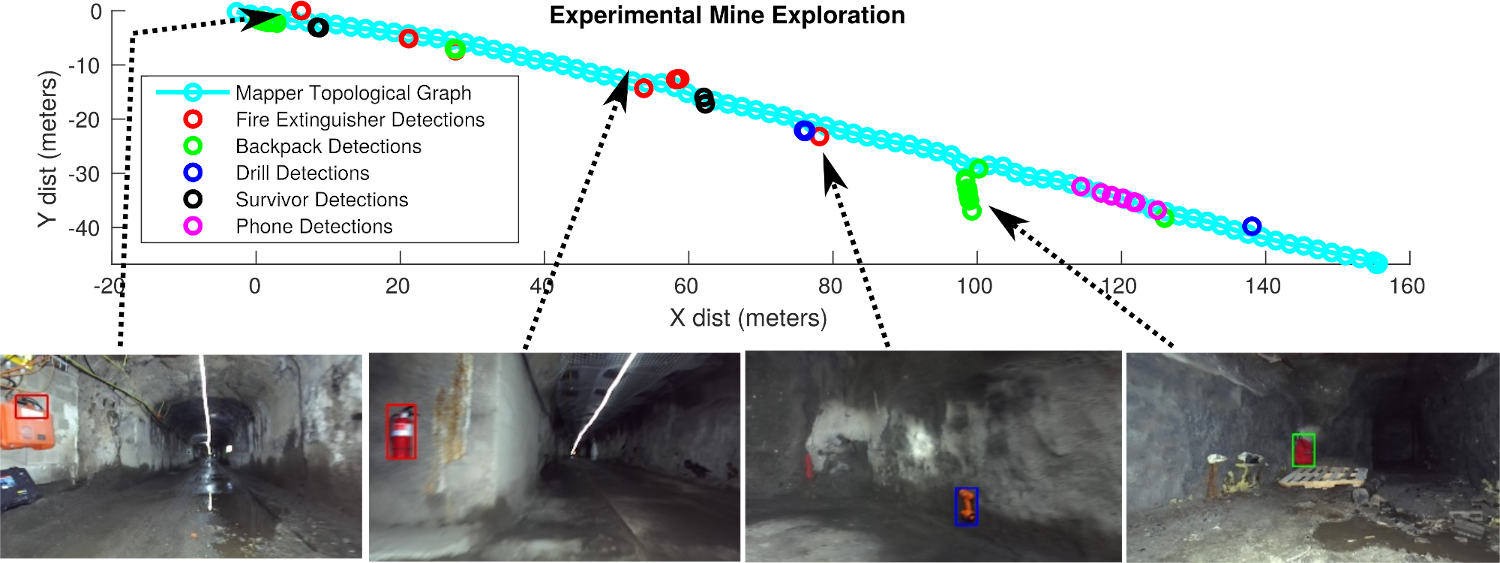}
    \caption{Traversal map showing 3 detected artifacts and a false positive.  Note also the cell phone detections in magenta.}
    \vspace{-1em}
    \label{fig:exp_mine}
\end{figure*}

\begin{figure}[ht]
    \centering
    \includegraphics[width=0.5\textwidth]{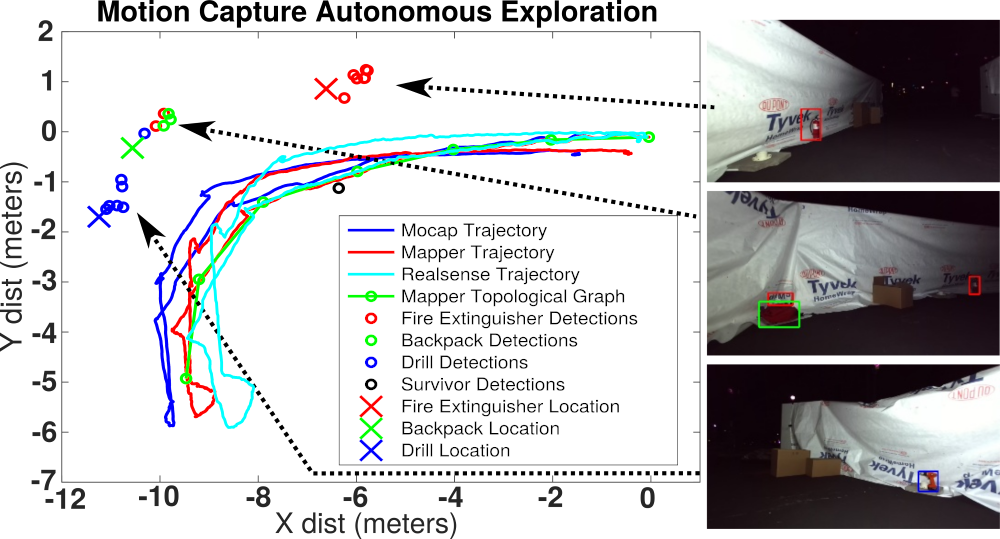}
    \caption{ Robot trajectories estimated from onboard mapper and RealSense compared with ground truth (mocap) for an autonomous there-and-back mission.}
    \vspace{-1em}
    \label{fig:mocap_exp}
\end{figure}

\begin{figure}[ht]
    \centering
    \includegraphics[width=0.5\textwidth]{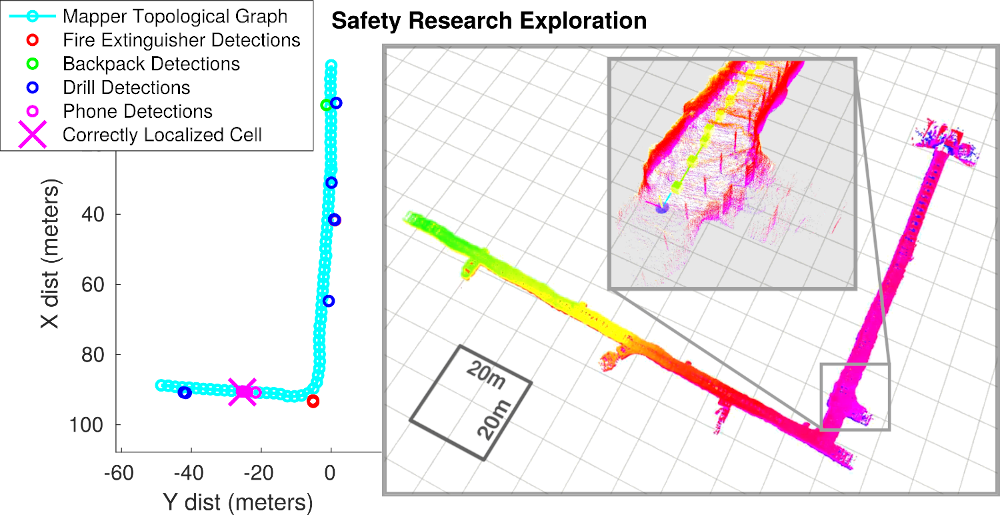}
    \caption{Traversal map as well as reconstructed pointcloud.  Note the correctly detected cell phone \textasciitilde 100m in to the mine.  The topological graph ends where the robot left communication range of the base station.}
    \vspace{-2em}
    \label{fig:sr_mine}
\end{figure}

\begin{table}[ht]
    \centering
    \begin{tabular}{c}
    \begin{tabular}{c|c|c}
         Artifact & Err (m) & Human Reviewed Err (m) \\ \hline
         Fire Extinguisher & 1.40 & 0.81 \\
         Backpack & 0.92 & 0.92 \\
         Drill & 5.83 & 0.55
    \end{tabular}
    \\
    \\
    \begin{tabular}{c|c|c}
         Sensor & Avg Error (m) & Std Dev Error (m) \\ \hline
         Ours (ATE) & 0.75 & 0.39 \\
         Realsense (ATE) & 1.12 & 0.61 \\
         Ours (RPE, 2 sec) & 0.080 & 0.057 \\
         Realsense (RPE, 2 sec) & 0.080 & 0.072
    \end{tabular}
    \end{tabular}
    \caption{Errors for artifact localization and pose for lab experiment.  Artifact errors are given for the raw data as well as data with misclassifications removed.  RPE is computed with a fixed delta of 2 seconds.}
    \label{tab:error_stats}
    \vspace{-1em}
\end{table}

In order to validate our approach in controlled circumstances, we performed a short autonomous mission in an outdoor motion capture system to provide ground truth estimates of the robot's pose using a simulated tunnel and a short mission time-out.  We additionally set up a drill, backpack, and fire extinguisher in the environment.  Fig.~\ref{fig:mocap_exp} shows the overlaid mapper and ground truth (motion capture) trajectories as well as the detection locations against ground truth.  It also shows the trajectory from the Realsense tracking camera.  All objects are localized well within a meter, and the total error after a \textasciitilde 30m traversal was on the order of a meter, giving the mapper \textasciitilde 3\% error.  Some false positives are shown, such as the backpack being incorrectly identified as a fire extinguisher in some frames, but these cases are easily filtered out by the human operator.  We additionally show numerical results in Table \ref{tab:error_stats}, noting that after operator filtering, the mean artifact errors are within a meter, as is the absolute trajectory error (ATE).  Relative pose error (RPE) is additionally computed with a fixed moving window of 2 seconds, and both show our LiDAR-based mapping outperform the visual-inertial RealSense tracking.

\subsection{Mine Experiments}

We additionally performed a number of tests as part of the DARPA SubT Tunnel Circuit at the NIOSH Research mine.  Over the course of 10-15 missions over 4 days, there were 3 primary failure modes.  About half of the missions suffered some sort of sensory failure, typically either Ouster LiDAR shutdown (likely due to a power supply problem) or RealSense failure (tracking completely diverging).  Robots typically required multiple startups before all sensors would start correctly, highlighting the usefulness of the launch manager's monitoring capability for rapid debugging under time constraints in the field.  All long duration missions, such as those shown in Figs.~\ref{fig:sr_mine} and \ref{fig:exp_mine}, ended with the robot falling on gravel, slippery surfaces, or dips in the surface.  The high level tunnel tracking worked very well, but the local planner tended to be overconfident, and the simple traversability metric of gradient did not well-capture the subtlety of different surfaces (e.g. gravel or wet metal) being dangerous.  This problem well captures the key flaw of a weight-based continuum planner.  If a dangerous path is sufficiently shorter than a more lengthy, safe, path, it can still be the cheaper path overall.  This problem was further exacerbated by our planning on a cartesian grid.  It was inherently much more expensive for the robot to travel across the grid diagonally as the planner would have to zig-zag along the grid before pruning occurred.  A motion-primitive based planner, vision-based traversability analysis, and hard limits on traversable terrain would all likely improve performance and better utilize the theoretical mobility benefit of legged robots.  Nonetheless, given that each step the robot takes is approximately 10cm, a 165m traverse as in Fig.~\ref{fig:exp_mine} corresponds to over a thousand successful steps taken by the planner.  In the Safety Research Mine (Fig. ~\ref{fig:sr_mine}), the robot travelled 240m before slipping and falling on gravel, corresponding to over two thousand steps, some of which were over significant obstacles such as train tracks.

In all runs, the detector had a 100\% true positive rate.  In other words, human analysis of the imagery did not reveal any artifact that the detector did not.  In the Experimental Mine (Fig.~\ref{fig:exp_mine}), 4 artifacts were correctly detected over the course of a 165m traversal as well as some false positives.  These extra detections were easily filtered out at the basestation by the human operator.  In addition, the distributed database communicated all detections back to the operator as expected as well as the depth panoramas and pose graph needed to reconstruct the global pointcloud such as that shown in Fig.~\ref{fig:sr_mine}.  However, at this mine, we were unable to localize these artifacts within 5 meters due to the spatial smoothness of the mine corridor and our reliance on LiDAR for localization.  We are currently exploring methods for better incorporating optical information to the mapping system to improve performance.  In the Safety Research Mine (Fig.~\ref{fig:sr_mine}), after \textasciitilde 100m, a cell phone was correctly detected and reported to within 5m, corresponding to less than 5\% error.
\section{Conclusion}
In this work, we have presented the exploration, planning, mapping, and detection technologies to autonomously explore, map, and detect artifacts in underground tunnel environments using multiple legged robots.  We have additionally demonstrated the feasibility and functionality of such a system in controlled lab experiments as well as field testing.  We note that our mapping, detection, and communication algorithms can also be applied to a more general class of multi-robot exploration problems.  Finally, we have identified areas for further work in mapping and control to increase robustness and accuracy.


\bibliographystyle{bib/IEEEtran}
\bibliography{bib/bib.bib}

\end{document}